\documentclass{bmvc2k}

\title{Colour Constancy: Biologically-inspired Contrast Variant Pooling Mechanism}

\addauthor{Arash Akbarinia}{www.cvc.uab.es/people/sakbarinia/}{1}
\addauthor{Raquel Gil Rodr\'{i}guez}{http://ip4ec.upf.edu/user/53}{2}
\addauthor{C. Alejandro Parraga}{www.cvc.uab.es/people/aparraga/}{1}

\addinstitution{
 Centre de Visi\'{o} per Computador\\
 Universitat Aut\`{o}noma de Barcelona\\
 Barcelona, Spain
}
\addinstitution{
 Department of Information and Communication Technologies\\
 Universitat Pompeu Fabra\\
 Barcelona, Spain
}

\runninghead{Arash Akbarinia}{Contrast Variant Pooling Mechanism}

\def\eg{\emph{e.g}\bmvaOneDot}
\def\ie{\emph{i.e}\bmvaOneDot}

\DeclareMathOperator*{\argmax}{arg\,max}

\usepackage{multirow,bigstrut}
\usepackage{tabu}
\usepackage[percent]{overpic}
\usepackage{tabularx,booktabs}

\begin{document}

\maketitle

\begin{abstract}

Pooling is a ubiquitous operation in image processing algorithms that allows for higher-level processes to collect relevant low-level features from a region of interest. Currently, max-pooling is one of the most commonly used operators in the computational literature. However, it can lack robustness to outliers due to the fact that it relies merely on the peak of a function. Pooling mechanisms are also present in the primate visual cortex where neurons of higher cortical areas pool signals from lower ones. The receptive fields of these neurons have been shown to vary according to the contrast by aggregating signals over a larger region in the presence of low contrast stimuli. We hypothesise that this contrast-variant-pooling mechanism can address some of the shortcomings of max-pooling. We modelled this contrast variation through a histogram clipping in which the percentage of pooled signal is inversely proportional to the local contrast of an image. We tested our hypothesis by applying it to the phenomenon of colour constancy where a number of popular algorithms utilise a max-pooling step (\eg White-Patch, Grey-Edge and Double-Opponency). For each of these methods, we investigated the consequences of replacing their original max-pooling by the proposed contrast-variant-pooling. Our experiments on three colour constancy benchmark datasets suggest that previous results can significantly improve by adopting a contrast-variant-pooling mechanism.

\end{abstract}

\section{Introduction}
\label{sec:intro}

Many computer vision frameworks contain a pooling stage that combines local responses at different spatial locations~\cite{boureau2010theoretical}. This operation is often a sum- or max-pooling mechanism implemented by a wide range of algorithms, \eg in feature descriptors (such as SIFT~\cite{lowe2004distinctive} and HOG~\cite{dalal2005histograms}) or convolutional neural networks~\cite{lecun1990handwritten,scherer2010evaluation}. Choosing the correct pooling operator can make a great difference in the performance of a method~\cite{boureau2010theoretical}. Current standard pooling mechanisms lack the desired generalisation to find equilibrium between frequently-occurring and rare but informative descriptors~\cite{murray2014generalized}. Therefore, many computer vision applications can benefit from a more dynamic pooling solution that takes into account the content of pooled signals.

Such pooling operators are commonly used in modelling the phenomenon of colour constancy (\ie a visual effect that makes the perceived colour of surfaces remain approximately constant under changes in illumination~\cite{foster2011color,hubel2000perception})  both in biological~\cite{carandini2012normalization} and computational~\cite{retinax} solutions. Despite decades of research, colour constancy still remains as an open question~\cite{foster2011color}, and solutions to this phenomenon are important from a practical point of view, \eg camera manufacturers need to produce images of objects that appear the same as the actual objects in order to satisfy customers. Motivated by above, in this article we propose a contrast-variant-pooling mechanism and investigate its feasibility in the context of computational colour constancy.



\subsection{Computational Models of Colour Constancy}

Mathematically, the recovery of spectral reflectance from a scene illuminated by light of unknown spectral irradiance is an ill-posed problem (it has infinite possible solutions). The simplest and  most popular solution has been to impose some arbitrary assumptions regarding the scene illuminant or its chromatic content. Broadly speaking colour constancy algorithms can be divided into two categories: (i) low-level driven, which reduce the problem to solving a set of non-linear mathematical equations~\cite{retinax,vandeWeijerTIP2007,7018983}, and (ii) learning-based, which train machine learning techniques on relevant image features~\cite{forsyth1990novel,funt2004estimating,agarwal2007machine}. Learning-based approaches may offer the highest performance results, however, they have major setbacks which make them unsuitable in certain conditions: (a) they rely heavily on training data that is not easy to obtain for all possible situations, and (b) they are likely to be slow~\cite{gijsenij2011color} and unsuitable for deployment inside limited hardware. A large portion of low-level driven models can be summarised using a general Minkowski expression \cite{finlayson2004shades,vandeWeijerTIP2007} such as:
\begin{equation}
L_c(p) = \left( \int_{\Omega} \left[ f_c(x) \right]^{p} dx \right)^{\frac{1}{p}} = ke_c ,
\label{eq:mink}
\end{equation}
where $L$ represents the illuminant at colour channel $c \in \lbrace R,G,B\rbrace$; $f(x)$ is the image's pixel value at the spatial coordinate $x\in \Omega$; $p$ is the Minkowski norm; and $k$ is a multiplicative constant chosen such that the illuminant colour, $e$, is a unit vector. 

Distinct values of the Minkowski norm $p$ results into different pooling mechanisms. Setting $p = 1$ in Eq.~\ref{eq:mink} reproduces the well known Grey-World algorithm (\ie sum-pooling), in which it is assumed that all colours in the scene average to grey~\cite{buchsbaum1980spatial}. Setting $p = \infty$ replicates the White-Patch algorithm (\ie max-pooling), which assumes the brightest patch in the image corresponds to the scene illuminant~\cite{retinax}. In general, it is challenging to automatically tune $p$ for every image and dataset. At the same time inaccurate $p$ values may corrupt the results noticeably~\cite{gijsenij2011color}.

The Minkowski framework can be generalised further by replacing $f(x)$ in Eq.~\ref{eq:mink} with its higher-order derivatives~\cite{vandeWeijerTIP2007}. These non-linear solutions are analogous to centre-surround mechanisms of visual cortex~\cite{land1986alternative}, which is typically modelled by a Difference-of-Gaussians (DoG) operators where a narrower, positive Gaussian plays the role of the ``centre'' and a broader, negative Gaussian  plays the role of the ``surround''~\cite{enroth1966contrast,marr1980theory}. Recently, a few biologically-inspired models of colour constancy grounded on DoG have offered promising results~\cite{7018983,parraga2016colour}, however their efficiency largely depends on finding an optimum pooling strategy for higher cortical areas. In short, pooling is a crucial component of many colour constancy models driven by low-level features (or even in deep-learning solutions~\cite{barron2015convolutional,fourure2016mixed}). In the primate visual systems the size of the receptive field varies according to the local contrast of the light falling on it~\cite{Shushruth2069,angelucci2013beyond} presenting a dynamic solution dependent on the region of interest.



The low-level models mentioned above are related to the early stages of visual processing, \ie the primary visual cortex (area V1), that are likely to be involved in colour constancy. Physiological measures suggest that although receptive fields triple in size from area V1 to area V2~\cite{wilson2014configural}, their basic circuity with respect to surround modulation is similar, \ie keeping the same size dependency with contrast properties found in V1~\cite{Shushruth2069}. This is consistent with the large body of physiological and psychophysical literature highlighting the significance of contrast in the visual cortex. In computer vision,  contrast-dependent models have also shown encouraging results in various applications such as visual attention~\cite{itti2001computational}, tone mapping~\cite{reinhard2002photographic}, and boundary detection~\cite{BMVC2016_12,akbarinia2017feedback}, to name a few. From these we can hypothesise the convenience and explanatory value of various ``pooling strategies'' such as those proposed by previous colour constancy methods. In the rest of this work we will explore the advantages of replacing the different feed-forward (pooling) configurations of some successful colour constancy models~\cite{retinax,vandeWeijerTIP2007,7018983} by that of the visual system (as described by current neurophysiological literature \cite{Shushruth2069,angelucci2013beyond}). Our aim in doing so is dual, on the one hand we want to explore the technological possibilities of creating a more efficient algorithm and on the other hand we would like to test the idea that contrast-variant-pooling might play an important role in colour constancy.



\subsection{Summary of the Contributions}

\begin{figure}[ht]
\centering
   \includegraphics[width=\linewidth]{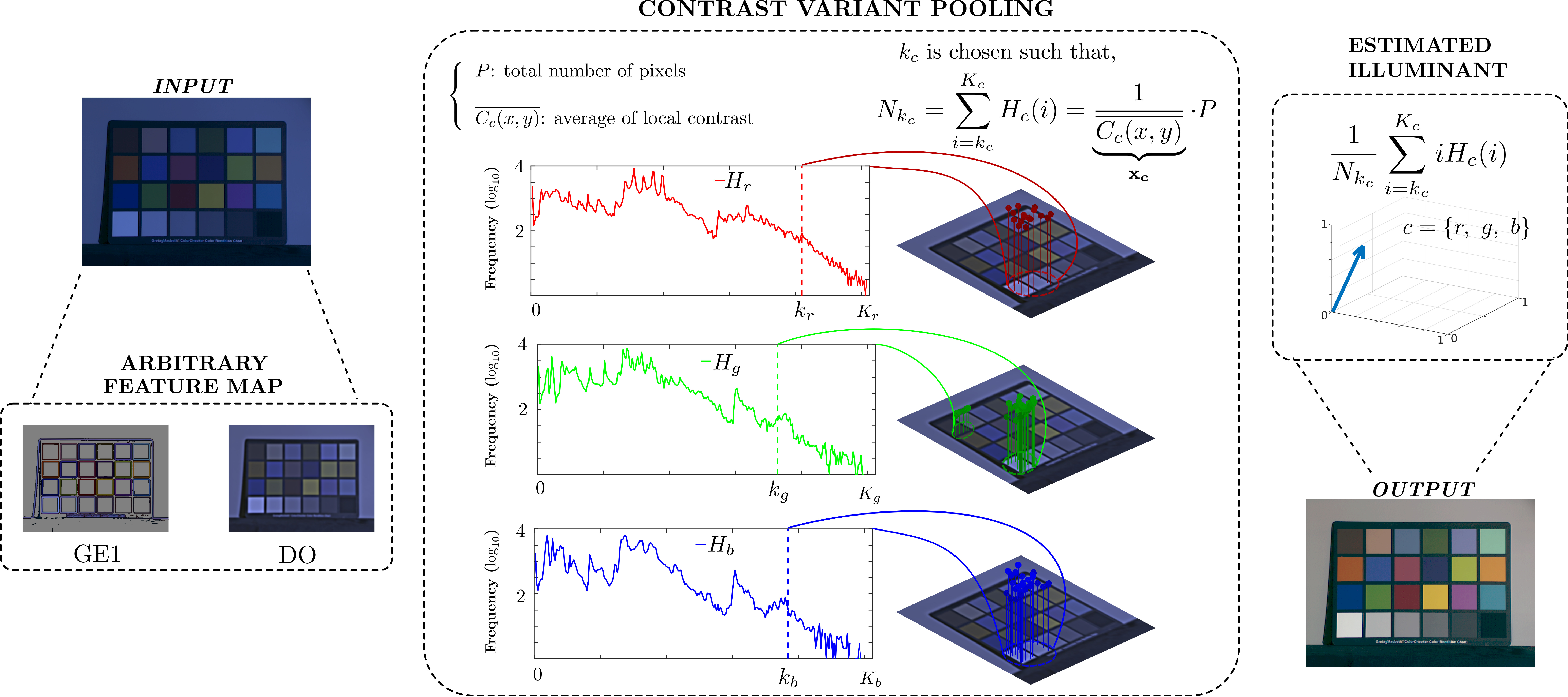}
   \caption{Flowchart of the proposed contrast-variant-pooling (CVP) mechanism in the context of colour constancy. We implemented CVP through a \textit{top-x-percentage-pooling}. Given an input image or a feature map: (i) value of the \textit{x} is computed according to the inverse of local contrast for each channel, and (ii) we estimate the scene illuminant as the average value of the \textit{top-x-percentage} of pooled pixels (those on the right side of the depicted dashed lines).}
\label{fig:v4pooling}
\end{figure}

In the present article we propose a generic contrast-variant-pooling (CVP) mechanism that can replace standard sum- and max-pooling operators in a wide range of computer vision applications. Figure~\ref{fig:v4pooling} illustrates the flowchart of CVP, which is based on local contrast and therefore it offers a dynamic solution that adapts the pooling mechanism according to the content of region of interest. We tested the feasibility of CVP in the context of colour constancy by substituting the pooling operation of four algorithms: (a) White-Patch~\cite{retinax}, (b) first-order Grey-Edge~\cite{vandeWeijerTIP2007}, (c) second-order Grey-Edge~\cite{vandeWeijerTIP2007}, and (d) Double-Opponency~\cite{7018983} in three benchmark datasets. Results of our experiments show the quantitative and qualitative benefits of CVP over max-pooling.

\section{Model}
\label{sec:method}

\subsection{Max-pooling Colour Constancy}
One of the earliest computational models of colour constancy (White-Patch) is grounded on the assumption that the brightest pixel in an image corresponds to a bright spot or specular reflection containing all necessary information about the scene illuminant \cite{retinax}. Mathematically this is equivalent to a max-pooling operation over the intensity of all pixels:
\begin{equation}
  L_c = \argmax_{x,y} I_c(x,y),
\end{equation}
where $L$ represents the estimated illuminant at each chromatic channel $c \in \left\lbrace R, G, B \right\rbrace$; $I$ is the original image and $(x,y)$ are the spatial coordinates in the image domain.

One important flaw of this simple approach is that a single bright pixel can misrepresent the whole illuminant. Furthermore, the White-Patch algorithm may fail in the presence of noise or clipped pixels in the image due to the limitations of the max-pooling operator \cite{funt1998machine}. One approach to address these issues is to account for a larger set of ``white'' points by pooling a small percentage of the brightest pixels (\eg the top $1\%$) \cite{ebner2007color},  an operation referred as \textit{top-x-percentage-pooling}. In this manner, the pooling mechanism is collectively computed considering a group of pixels rather than a single one. This small variant might be a crucial factor in the estimation of the scene illuminant \cite{joze2012role}. A similar mechanism has also been deployed successfully in other applications such as shadow removal~\cite{finlayson2002removing}.

In practice, given the chosen \textit{x-percentage}, the \textit{top-x-percentage-pooling} can be implemented through a histogram-based clipping mechanism~\cite{finlayson2002removing,ebner2007color}. Let $H$ be the histogram of the input image $I$, and let $H_c(i)$ represents number of pixels in colour channel $c$ with intensity $i \in [0, \cdots, K_c]$ (histogram's domain). The scene illuminant $L_c$ is computed as:
\begin{equation}
L_c = \frac{1}{N_{k_c}} \sum_{i=k_c}^{K_c} i \cdot H_c(i),
\label{eq:illuminant}
\end{equation}
where $N_{k_c}$ is the total number of pixels within the intensity range $k_c$ to $K_c$. The values of $k_{c= \{ R, G, B \}}$ are determined from the chosen \textit{x-percentage} such that:
\begin{equation}
N_{k_c} =  \sum_{i=k_c}^{K_c} H_c(i) = x_c \cdot P,
\label{eq:hist}
\end{equation}
where $P$ is the total number of pixels in the image and $x_c$ is the chosen percentage for colour channel $c$. Within this formulation it is very cumbersome to define a universal, optimally-fixed percentage of pixels to be pooled \cite{ebner2007color} and consequently the free variable $x$ requires specific tuning for each image or dataset individually. Naturally, this limitation restricts the usability of the \textit{top-x-percentage-pooling} operator. In the following sections we show how to automatically compute this percentage, based on the local contrast of each image.

\subsection{Pooling Mechanisms in the Visual Cortex}
We know from the physiology of the cerebral cortex that neurons in higher cortical areas pool information from lower areas over increasingly larger image regions. Although the exact pooling mechanism is yet to be discovered, ``winner-takes-all'' and ``sparse coding'' kurtotical behaviour are common to many groups of neurons all over the visual cortex~\cite{carandini2012normalization,olshausen1996emergence}, and it is conceivable that a mechanism analogous to max-pooling might be present within the cortical layers. Indeed such behaviour has been discovered across a population of cells in the cat visual cortex \cite{lampl2004intracellular} and the activation level of cells with max-like behaviour was reported to vary depending on the contrast of visual stimuli.

Results reported by~\cite{lampl2004intracellular} suggest an inverse relationship between the contrast of a stimulus and the percentage of the signal pooled. When pooling neurons were exposed to low contrast stimuli, their responses shifted slightly away from pure max-pooling (selecting the highest activation response within a region) towards integrating over a larger number of highly activated neurons. In the language of computer vision, this can be regarded as \textit{top-x-percentage-pooling}, where \textit{x} assumes a smaller value in high contrast and a larger value in low contrast. Interestingly, the pooling of those neurons remained always much closer to max-pooling than to the linear integration of all neurons (sum-pooling)~\cite{lampl2004intracellular}. Mathematically, this can be interpreted as having a very small  (\textit{top-x-percentage-pooling}) $x$  value. 

It does not come as a great surprise that the pooling mechanism in the visual cortex depends on the stimulus' contrast. There is a large body of physiological studies showing that receptive fields (RF) of neurons are contrast variant (for a comprehensive review refer to~\cite{angelucci2013beyond}). Quantitative results suggest that RFs in visual area one (V1) of awake macaques double their size when measured at low contrast~\cite{Shushruth2069}. Similar expansions have also been reported for the RFs of neurons in extrastriate areas, such as V2 and V4. This implies that a typical neuron in higher cortical areas that normally pool responses from its preceding areas over about three neighbouring spatial locations \cite{wilson2014configural} can access a substantially larger region to pool from in the presence of low contrast stimuli. This is in line with the reported pooling mechanism in the cat visual cortex~\cite{lampl2004intracellular}.

\subsection{Contrast Variant Pooling}

In order to model this contrast-variant-pooling mechanism, we first computed local contrast $C$ of the input image $I$ at every pixel location by means of its local standard deviation defined as:
\begin{equation}
C_c(x, y; \sigma) = \sqrt{ \frac{1}{\# \mathcal{N}_{\sigma}(x,y)}  \sum_{ \substack{ (x', y')  \in \mathcal{N}_{\sigma}(x,y) }} \big( I_c(x', y') - \mu (x, y) \big) ^2  }
\label{eq:contrastcalculation}
\end{equation}
where $c$ indexes each colour channel; $(x,y)$ are the spatial coordinates of a pixel; $\mu$ is an isotropic average kernel with size $\sigma$; and $\mathcal{N}_{\sigma}(x,y)$ represents the neighbourhood centred on pixel $(x,y)$ of radius $\sigma$.

To simulate this inverse relation between stimulus contrast and percentage of signal pooled \cite{lampl2004intracellular} in the \textit{top-x-percentage-pooling} operator, we determined the percentage $x_c$ in Eq.~\ref{eq:hist} as the average of inverse local contrast signals:
\begin{equation}
  x_c = \frac{1}{P} \sum_{x,y}\frac{1}{C_c(x,y; \sigma)}, \ \forall (x,y) \in \Omega,
  \label{eq:pestimation}
\end{equation}
where $C_c$ is computed from Eq. \ref{eq:contrastcalculation}, and $\Omega$ is the spatial image domain. In this fashion, instead of defining a fix percentage of signal (pixels) to be pooled (as in~\cite{finlayson2002removing}), we chose an adaptive percentage according to the contrast of image. In terms of colour constancy, this effectively relates the number of pooled pixels to compute the scene illuminant to the average contrast of an image. We illustrated this mechanism of contrast-variant-pooling in the central panel of Figure~\ref{fig:v4pooling}, where red, green and blue signals correspond to the histogram of each chromatic channel. Pixels on the right side of the dashed lines ($k_c$) are pooled. In this example, contrast is higher for the red signal and therefore a smaller percentage of cells are pooled in the red channel.

Bearing in mind that ``contrast'' is just a fraction in the range $[0,1]$ -- with $0$ characterising an absolutely uniform area and $1$ representing points with the highest contrast, \eg edges -- we can predict that the percentage $x$ will be a very small number for natural images where homogeneous regions are likely to form the majority of the scene. Consequently in our \textit{top-x-percentage-pooling} operator we always pool a small percentage. This is in agreement with observations of~\cite{lampl2004intracellular} which indicate that such pooling mechanism is always much closer to max-pooling than to sum-pooling.

\subsection{Generalisation to Other Colour Constancy Models}

There is a number of colour constancy models in the literature which are driven by low-level features that require a pooling mechanism on top of their computed feature maps in order to estimate the scene illuminant. In the Double-Opponency~\cite{7018983} algorithm this feature map is computed by convolving a colour-opponent representation of the image with a DoG kernel followed by a  max-pooling operation. In the Grey-Edge~\cite{vandeWeijerTIP2007} model, higher-order derivatives of the image are calculated through its convolution with the first- and second-order derivative of a Gaussian kernel. This is complemented by a Minkowski summation, which oscillates between sum- and max-pooling depending on its norm.

Similar to the White-Patch algorithm, the pooling mechanism of these models can also be replaced with our \textit{top-x-percentage-pooling} operator, where \textit{x} is computed according to the local contrast of image as explained above. The only difference is that instead of pooling from an intensity image (as in case of the White-Patch algorithm), Double-Opponency and Grey-Edge pool over their respective feature maps. This means that Eq.~\ref{eq:illuminant} receives a feature map $M$ instead of an intensity image $I$ as input.

\section{Experiments and Results}
\label{sec:results}

In order to investigate the efficiency of our model, we applied the proposed contrast-variant-pooling (CVP) mechanism to four different colour constancy algorithms whose source code were publicly available: White-Patch~\cite{retinax}, first-order Grey-Edge~\cite{vandeWeijerTIP2007}, second-order Grey-Edge~\cite{vandeWeijerTIP2007}, and Double-Opponency~\cite{7018983}. We simply replaced their max-pooling operator with our proposed pooling mechanism. To evaluate each method we used the recovery angular error defined as:
\begin{equation}
  \epsilon^{\circ} \left( e_e, e_t \right) = \arccos \left(\frac{<e_e , e_t>}{\Vert e_e \Vert  \Vert e_t \Vert }\right) ,
\end{equation}
where $< . >$ represents the dot product of the estimated illuminant $e_e$ and the ground truth $e_t$; and $\Vert . \Vert $ stands for the Euclidean norm of a vector. It is worth mentioning that this error measure might not correspond precisely to observers' preferences~\cite{vazquez2009color}, however, it is the most commonly used comparative measure in the literature. We also computed the reproduction angular error~\cite{finlayson2014reproduction} in all  experiments (due to lack of space these results are not reported here). Readers are encouraged to check the accompanying  supplementary materials. 

We conducted experiments on three benchmark datasets\footnote{All source code and materials are available in the supplementary submission.}, (i) SFU Lab (321 images)~\cite{barnard2002data}, (ii) Colour Checker (568 images)~\cite{ColourCheckerDB}, and (iii) Grey Ball (11,346 images)~\cite{CiureaF03}. In Table~\ref{tab:resultssfulab} we have reported the best median and trimean angular errors for each of the considered methods (these metrics were proposed by~\cite{hordley2006reevaluation} and~\cite{gijsenij2009perceptual} respectively to evaluate colour constancy algorithms since they are robust to outliers). Mean angular errors are reported in the supplementary materials.

\begin{table}[ht]
\setlength{\tabcolsep}{4.7pt}
\centering
\begin{tabular}{@{\hskip 0.05in}ll@{\hskip 0.02in}|c|c|c|c|c|c|}
\cline{3-8}
 & & \multicolumn{2}{c|}{SFU Lab~\cite{barnard2002data}} & \multicolumn{2}{c|}{Colour Checker~\cite{ColourCheckerDB}} & \multicolumn{2}{c|}{Grey Ball~\cite{CiureaF03}} \\ \hline
\multicolumn{1}{|@{\hskip 0.05in}l@{\hskip 0.02in}}{Method} & & Median & Trimean & Median & Trimean & Median & Trimean \\
\cline{1-8}
\multicolumn{1}{|@{\hskip 0.05in}l@{\hskip 0.02in}}{\small White-Patch} & \cite{retinax} &  6.44 & 7.51 & 5.68 & 6.35 & 6.00 & 6.50  \\
\multicolumn{1}{|@{\hskip 0.05in}l@{\hskip 0.02in}}{\small Grey-Edge 1\textsuperscript{st}-order} & \cite{vandeWeijerTIP2007} & 3.52 & 4.02 & 3.72 & 4.76 & 5.01 & 5.80   \\
\multicolumn{1}{|@{\hskip 0.05in}l@{\hskip 0.02in}}{\small Grey-Edge 2\textsuperscript{nd}-order} & \cite{vandeWeijerTIP2007} & 3.22 & 3.65 & 4.27 & 5.19 & 5.72 & 6.39  \\
\multicolumn{1}{|@{\hskip 0.05in}l@{\hskip 0.02in}}{\small Double-Opponency} & \cite{7018983} & 2.37 & 3.27  & 3.46 & 4.38 & 4.62 & 5.28 \\
\hline \hline
\multicolumn{2}{|@{\hskip 0.05in}l@{\hskip 0.02in}|}{\textbf{\small CVP White-Patch}} & \textbf{2.99} & \textbf{3.42} & \textbf{2.97} & \textbf{3.45}  & \textbf{5.02} & \textbf{5.15}  \\
\multicolumn{2}{|@{\hskip 0.05in}l@{\hskip 0.02in}|}{\textbf{\small CVP Grey-Edge 1\textsuperscript{st}-order}} & \textbf{3.29} & \textbf{3.72} & \textbf{2.48} & \textbf{2.79}  & \textbf{4.70} & \textbf{5.17}  \\
\multicolumn{2}{|@{\hskip 0.05in}l@{\hskip 0.02in}|}{\textbf{\small CVP Grey-Edge 2\textsuperscript{nd}-order}} & \textbf{2.85} & \textbf{3.13} & \textbf{2.59} & \textbf{2.93}  & \textbf{4.65} & \textbf{5.05}  \\
\multicolumn{2}{|@{\hskip 0.05in}l@{\hskip 0.02in}|}{\textbf{\small CVP Double-Opponency}}& \textbf{2.02} & \textbf{2.56} & \textbf{2.39} & \textbf{2.84}  & \textbf{4.00} & \textbf{4.24}  \\
\hline
\end{tabular}
\caption{Recovery angular errors of four colour constancy methods under max- and contrast-variant-pooling (CVP) on three benchmark datasets. Lower figures indicate better performance.}
\label{tab:resultssfulab}
\end{table}

\begin{figure}[ht]
\centering
\begin{tabular}{@{\hskip 0.0in}c@{\hskip 0.04in}c@{\hskip 0.04in}c@{\hskip 0.04in}c@{\hskip 0.04in}c@{\hskip 0.04in}c@{\hskip 0.00in}}
\includegraphics[width=0.159\linewidth]{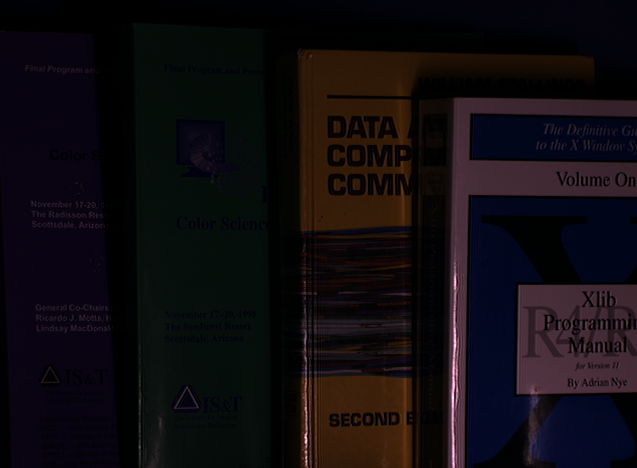} &
\includegraphics[width=0.159\linewidth]{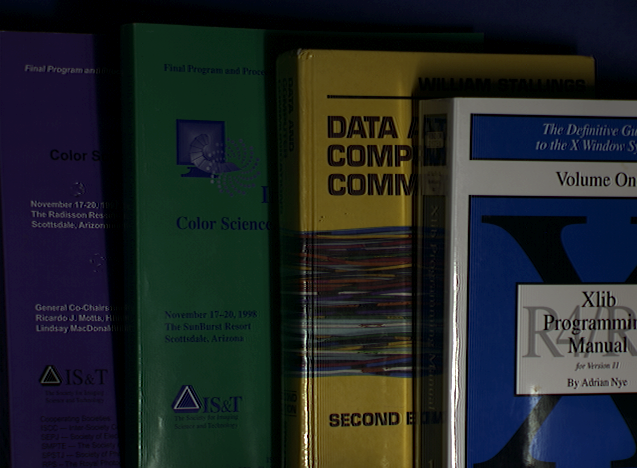} &
\begin{overpic}[width=0.159\linewidth]{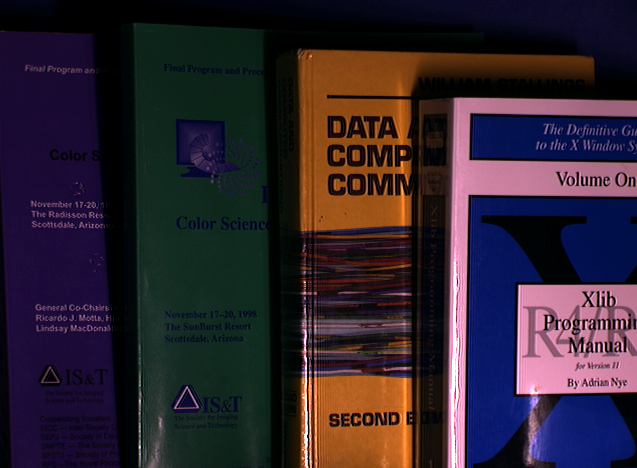}
  \put(68,6){\colorbox{white}{\parbox[c]{0.03\textwidth}{\textsc{\tiny{$9.73^\circ$}}}}}
\end{overpic} &
\begin{overpic}[width=0.159\linewidth]{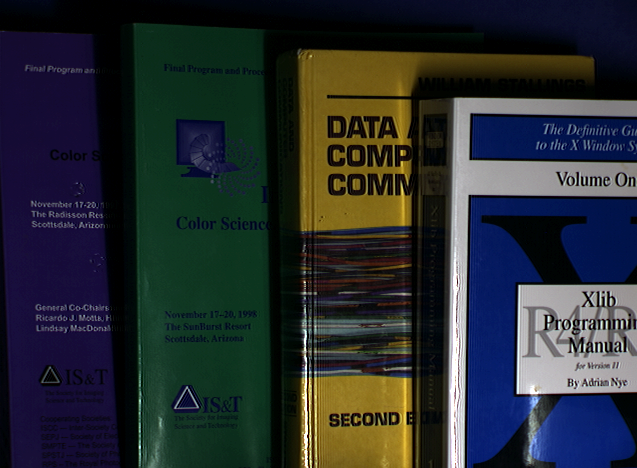}
  \put(68,6){\colorbox{white}{\parbox[c]{0.03\textwidth}{\textsc{\tiny{$1.83^\circ$}}}}}
\end{overpic} &
\begin{overpic}[width=0.159\linewidth]{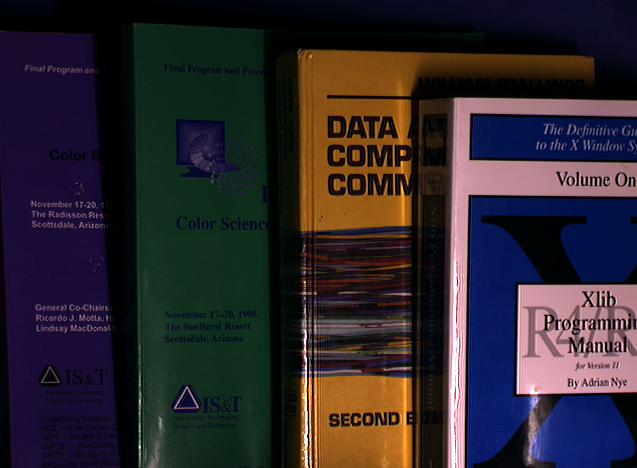}
  \put(68,6){\colorbox{white}{\parbox[c]{0.03\textwidth}{\textsc{\tiny{$6.73^\circ$}}}}}
\end{overpic} &
\begin{overpic}[width=0.159\linewidth]{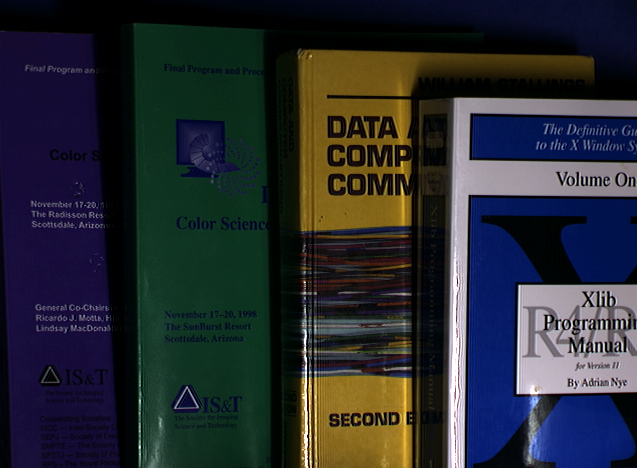}
  \put(68,6){\colorbox{white}{\parbox[c]{0.03\textwidth}{\textsc{\tiny{$2.59^\circ$}}}}}
\end{overpic} \\

\includegraphics[width=0.159\linewidth]{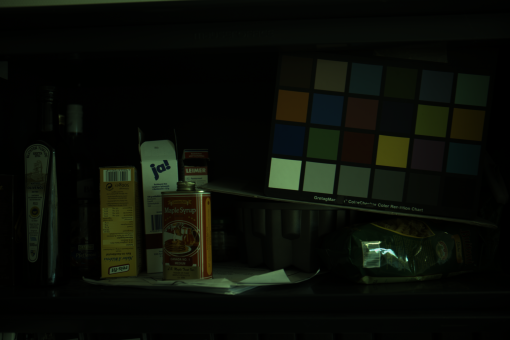} &
\includegraphics[width=0.159\linewidth]{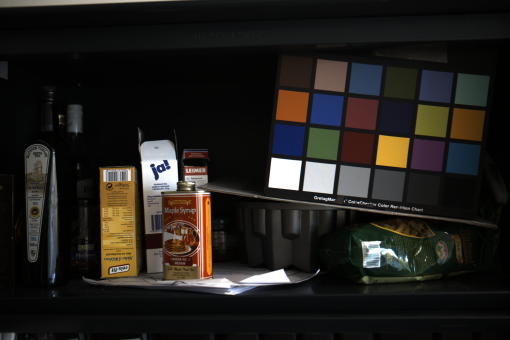} &
\begin{overpic}[width=0.159\linewidth]{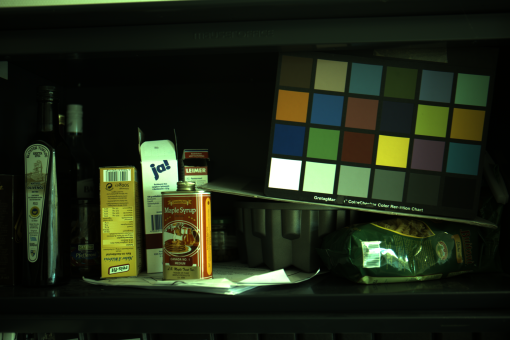}
  \put(68,6){\colorbox{white}{\parbox[c]{0.03\textwidth}{\textsc{\tiny{$11.39^\circ$}}}}}
\end{overpic} &
\begin{overpic}[width=0.159\linewidth]{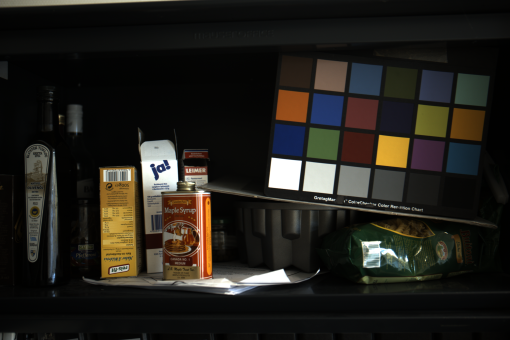}
  \put(68,6){\colorbox{white}{\parbox[c]{0.03\textwidth}{\textsc{\tiny{$1.52^\circ$}}}}}
\end{overpic} &
\begin{overpic}[width=0.159\linewidth]{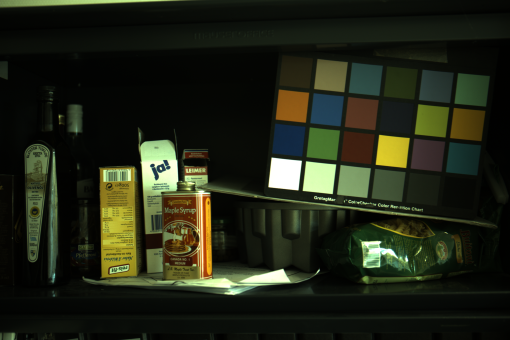}
  \put(68,6){\colorbox{white}{\parbox[c]{0.03\textwidth}{\textsc{\tiny{$10.31^\circ$}}}}}
\end{overpic} &
\begin{overpic}[width=0.159\linewidth]{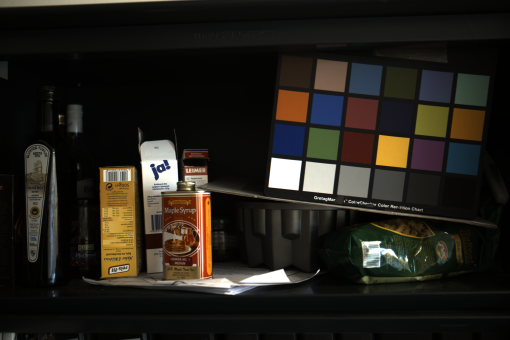}
  \put(68,6){\colorbox{white}{\parbox[c]{0.03\textwidth}{\textsc{\tiny{$2.34^\circ$}}}}}
\end{overpic} \\

\includegraphics[width=0.159\linewidth]{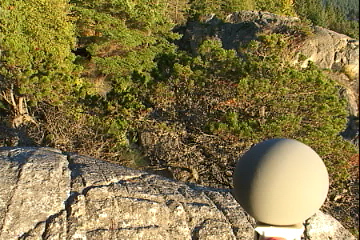} &
\includegraphics[width=0.159\linewidth]{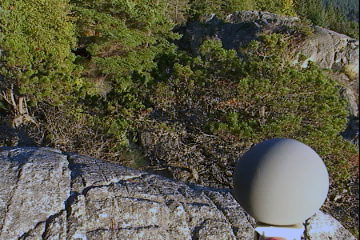} &
\begin{overpic}[width=0.159\linewidth]{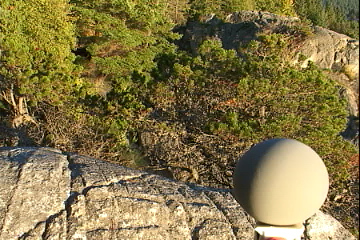}
  \put(68,6){\colorbox{white}{\parbox[c]{0.03\textwidth}{\textsc{\tiny{$8.45^\circ$}}}}}
\end{overpic} &
\begin{overpic}[width=0.159\linewidth]{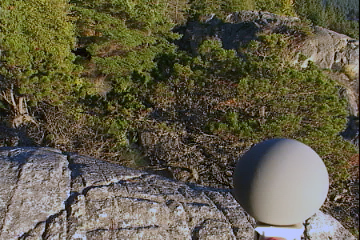}
  \put(68,6){\colorbox{white}{\parbox[c]{0.03\textwidth}{\textsc{\tiny{$1.45^\circ$}}}}}
\end{overpic} &
\begin{overpic}[width=0.159\linewidth]{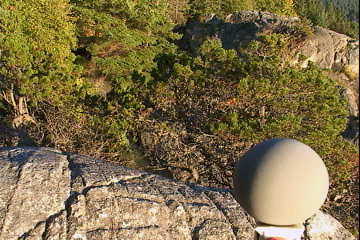}
  \put(68,6){\colorbox{white}{\parbox[c]{0.03\textwidth}{\textsc{\tiny{$7.18^\circ$}}}}}
\end{overpic} &
\begin{overpic}[width=0.159\linewidth]{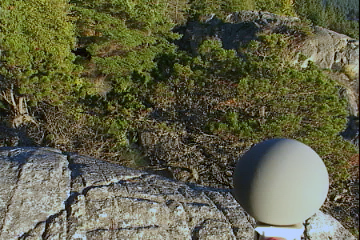}
  \put(68,6){\colorbox{white}{\parbox[c]{0.03\textwidth}{\textsc{\tiny{$1.69^\circ$}}}}}
\end{overpic} \\

Original & Ground Truth & WP & CVP WP & GE1 & CVP GE1
\end{tabular}
   \caption{Qualitative results of White-Patch (WP) and the first-order Grey-Edge (GE1) under max- and contrast-variant-pooling (CVP). Angular errors are indicated on the bottom right corner of each panel. Images are from the SFU Lab, Colour Checker and Grey Ball dataset respectively.}
\label{fig:qualres}
\end{figure}

Figure~\ref{fig:qualres} illustrates three exemplary results obtained by the proposed contrast-variant-pooling (CVP) operator for two colour constancy models: White-Patch and the first-order Grey-Edge. Qualitatively, we can  observe that CVP does a better job than max-pooling in estimating the scene illuminant. This is also confirmed quantitatively for the angular errors, shown at the right bottom side of each computed output.


\subsection{Influence of the Free Parameters}
\begin{figure}[ht]
\centering
\begin{tabular}{c c c}
\includegraphics[width=.3\linewidth]{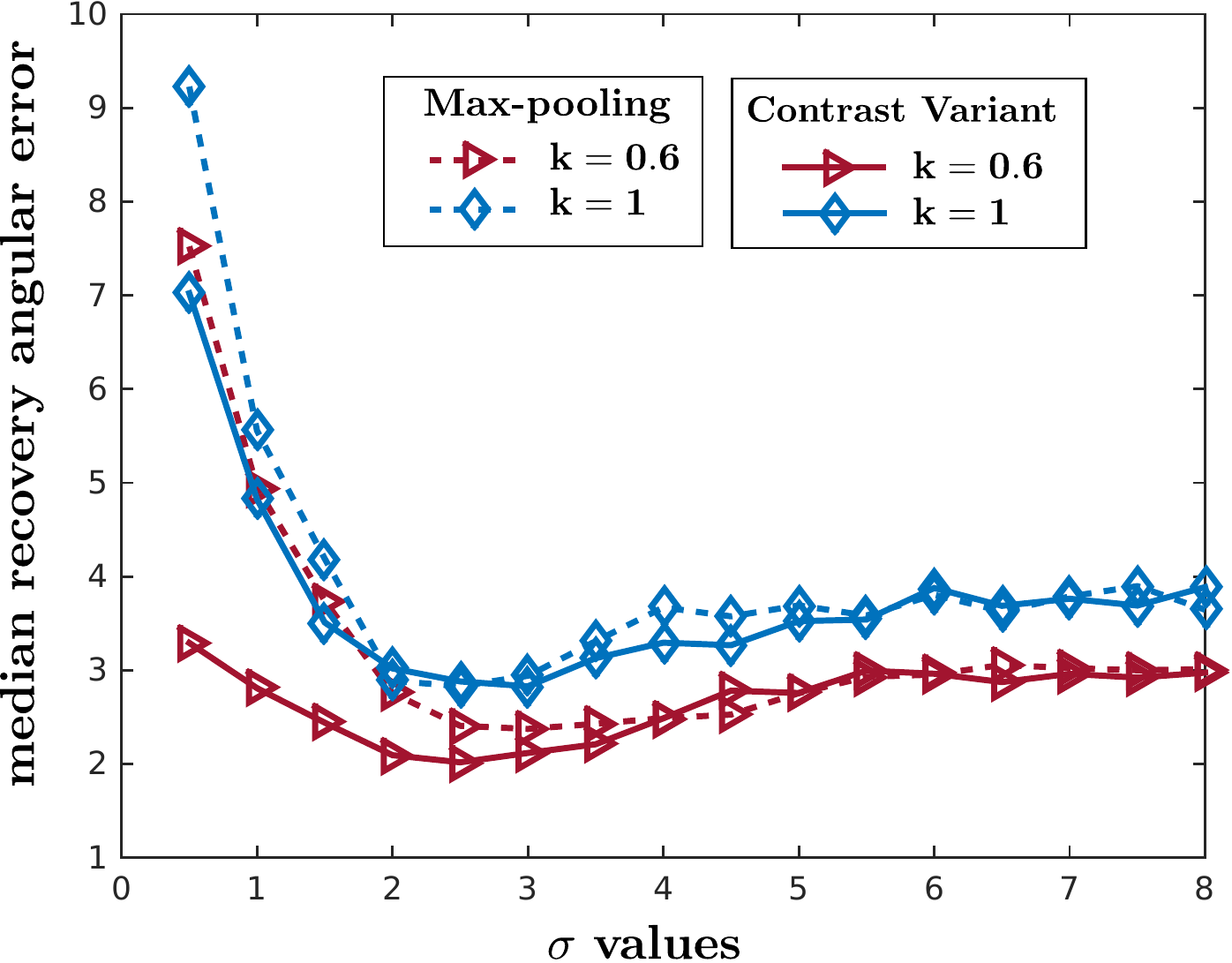} & 	\includegraphics[width=.3\linewidth]{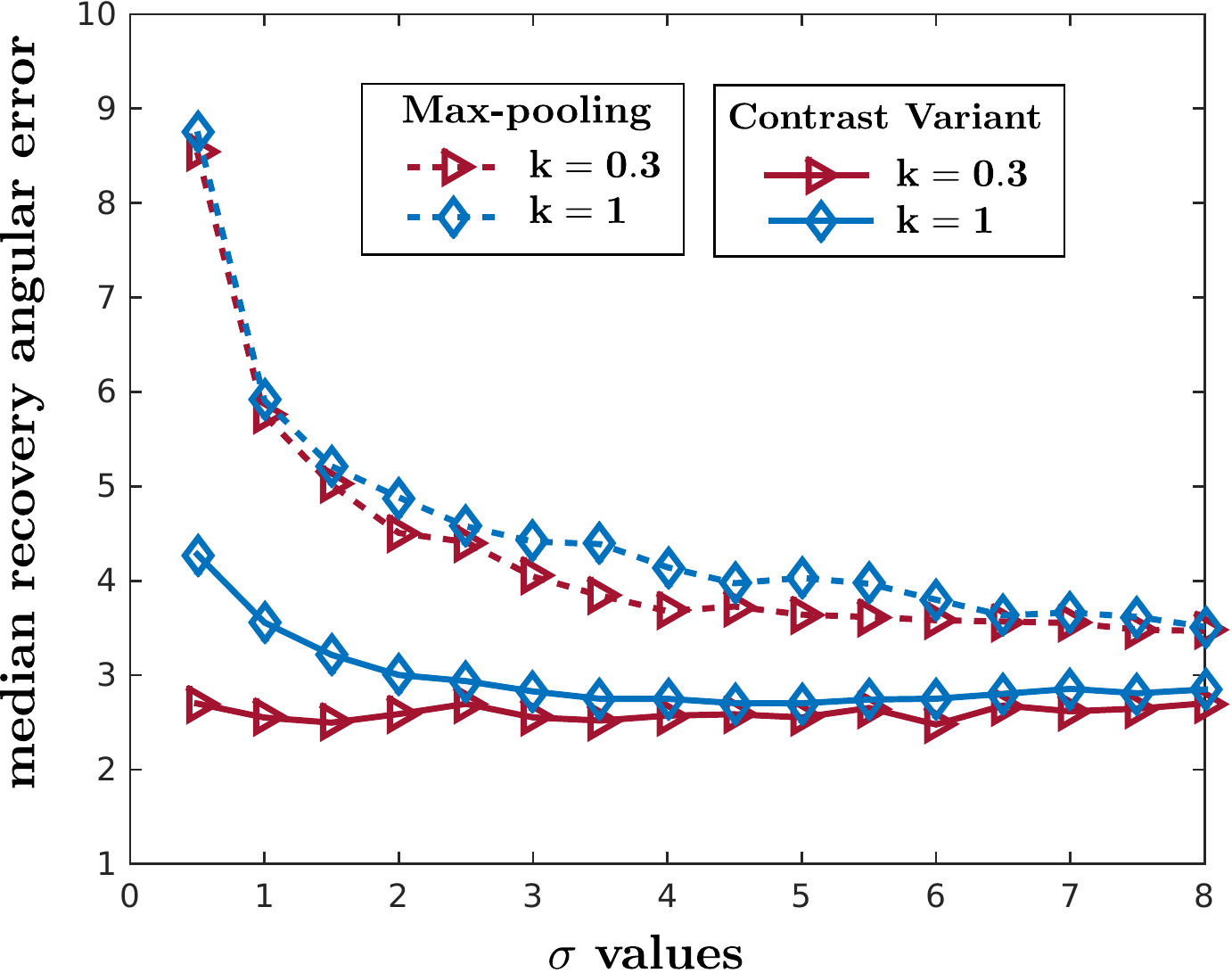} & \includegraphics[width=.3\linewidth]{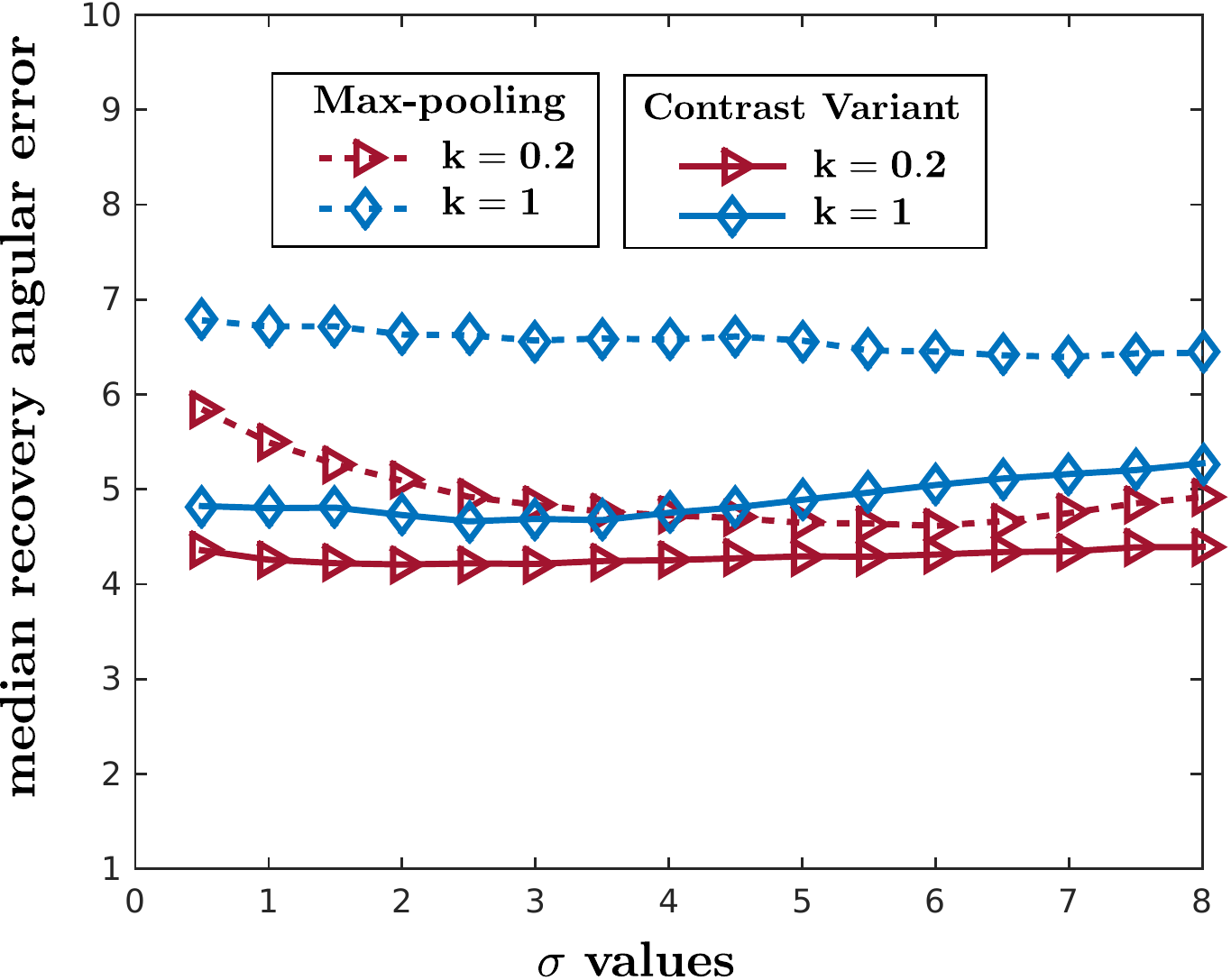} \\
SFU Lab~\cite{barnard2002data} & Colour Checker~\cite{ColourCheckerDB} & Grey Ball~\cite{CiureaF03}
\end{tabular}
\caption{The best and the worst results obtained by max- and contrast-variant-pooling for the free variables of Double-Opponency~\cite{7018983} algorithm ($k$ and $\sigma$).}
\label{results_do}
\end{figure}

For each free variable of the tested models we compared the performance of max- to contrast-variant-pooling. White-Patch does not have any free variable, therefore it is exempted from this analysis. In Figure~\ref{results_do} we have reported the impact of different $\sigma$s (receptive filed size) on Double-Opponency algorithm for the best and the worst results obtained by free variable $k$ in each dataset (results for all $k$s are available in the supplementary material). We can observe that almost in all cases contrast-variant-pooling outperforms max-pooling. The improvement is more tangible for the Colour Checker and Grey Ball datasets and in low $\sigma$s.

\begin{figure}[ht]
\centering
\begin{tabular}{c c c}
\includegraphics[width=.3\linewidth]{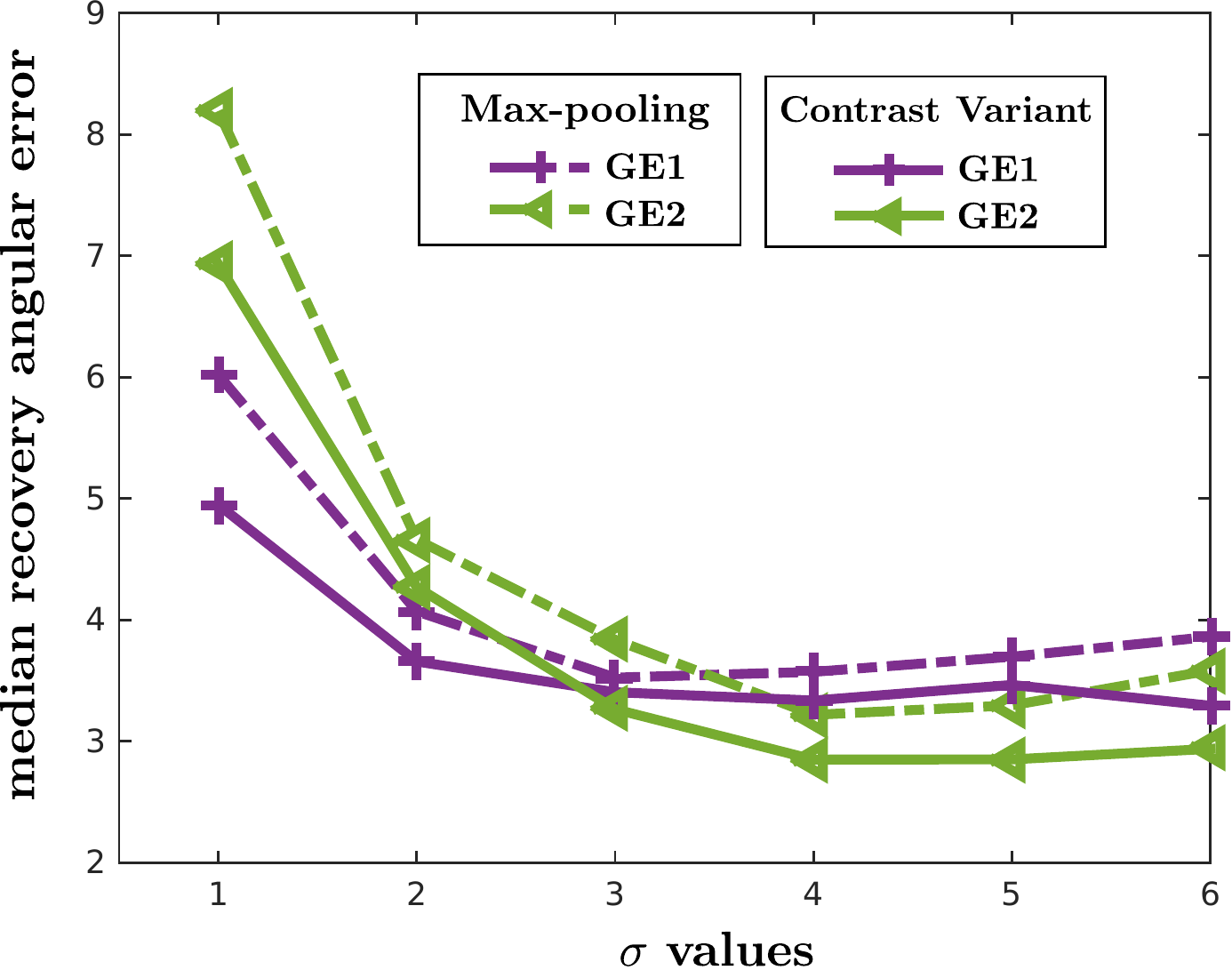} & 	\includegraphics[width=.3\linewidth]{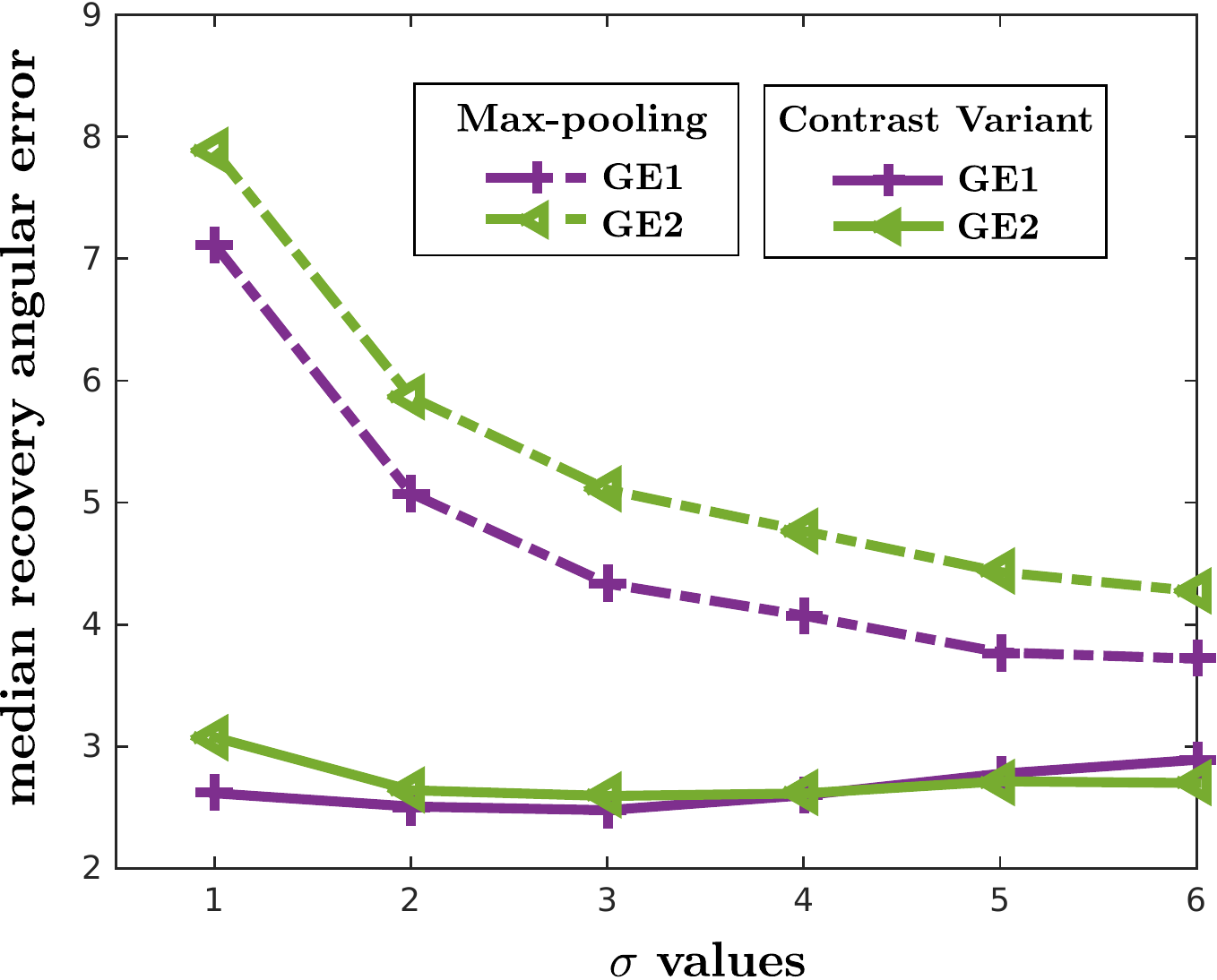} & \includegraphics[width=.3\linewidth]{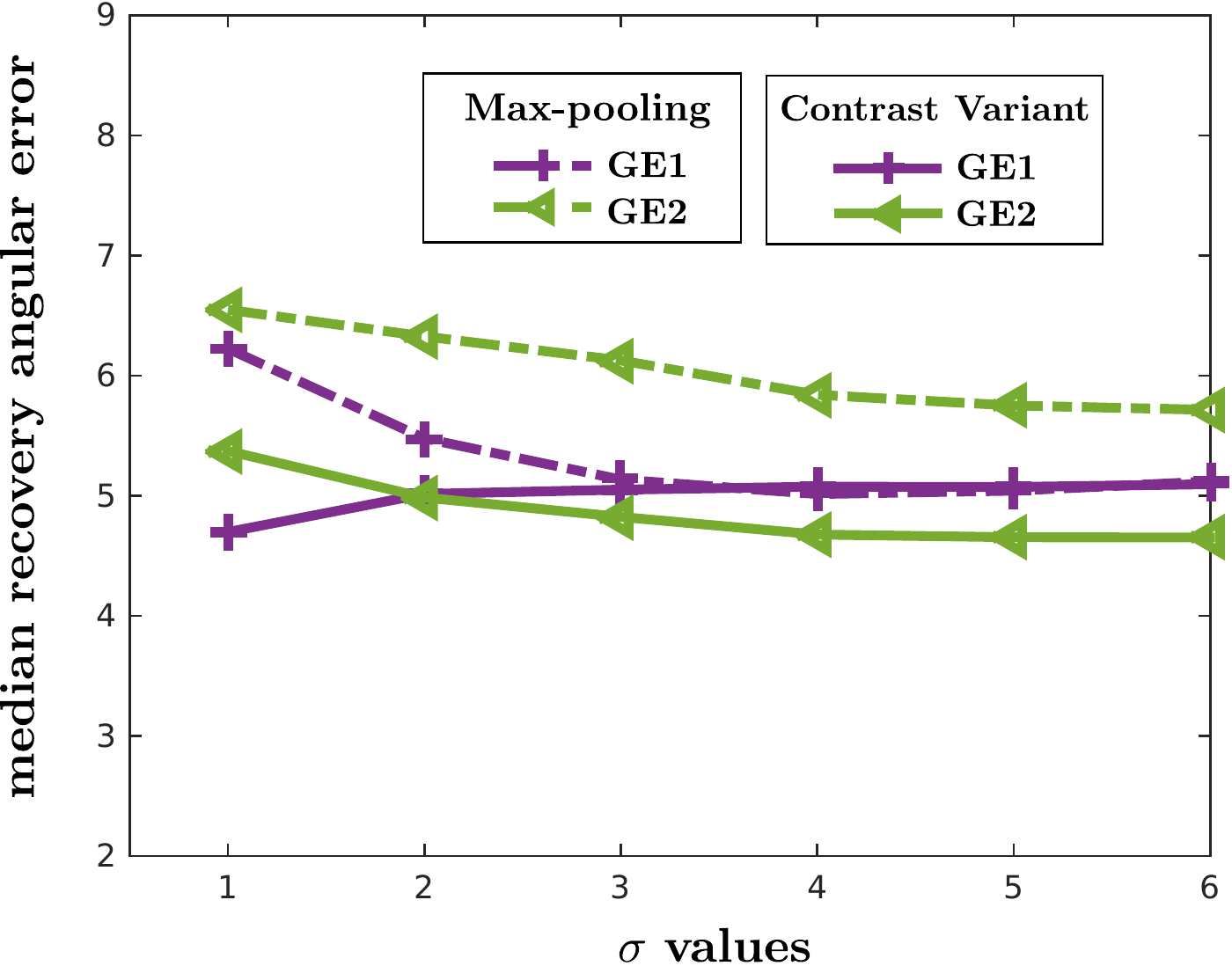} \\
SFU Lab~\cite{barnard2002data} & Colour Checker~\cite{ColourCheckerDB} & Grey Ball~\cite{CiureaF03}
\end{tabular}
\caption{Comparison of  max- and contrast-variant-pooling for the free variable $\sigma$ of Grey-Edge~\cite{vandeWeijerTIP2007} algorithm (both the first- and second-order derivatives).}
\label{results_ge}
\end{figure}

Figure~\ref{results_ge} illustrates the impact of different $\sigma$s (Gaussian size) on the first- and second-order Grey-Edge algorithm. We can observe similar patterns as with Double-Opponency (contrast-variant-pooling outperforms max-pooling practically in all cases). This improvement is more significant for low $\sigma$s, for the Colour Checker dataset and for the second-order derivative. It must be noted that the objective of this article was merely to study the performance of max-pooling and CVP on top of the Grey-Edge algorithm. However, the angular errors of our \textit{CVP Grey-Edge} happen to be in par with the best results reported for Grey-Edge (obtained by using the optimum Minkowski norm for each dataset~\cite{vandeWeijerTIP2007}), with the important caveat that CVP has no extra variables to be tuned, whereas in the Minkowski norm optimisation the value of $p$ must be hand-picked for each dataset.

From Figures~\ref{results_do} and~\ref{results_ge} we can observe that the greatest improvement occurs in the Colour Checker dataset. We speculate that one of the reasons for this is the larger range of intensity values in the Colour Checker dataset (16-bit) in comparison to the other two datasets that contain 8-bit images, therefore, an inaccurate max-pooling is more severely penalised.

\subsection{Discussion}
\label{sec:discussion}

We would like to emphasise that the objective of this article is not to improve state-of-the-art in colour constancy, but to show that contrast-variant-pooling (CVP) almost always produces improvements over max-pooling. Surprisingly, the results we obtained are even competitive with the state-of-the-art. For instance, in the SFU Lab dataset, the lowest reported angular error is 2.1 (obtained by an Intersection-based Gamut algorithm~\cite{barnard2000improvements}). This means that our \textit{CVP Double-Opponency} with an angular error of 2.0 outperforms the state-of-the-art in this dataset. In the Colour Checker and Grey Ball datasets there are a few learning-based models (\eg Exemplar-based method \cite{6588227}) that obtain lower angular errors in comparison to \textit{CVP Double-Opponency}, nevertheless our results are comparable with theirs. 

Physiological evidence besides, the better performance of CVP can be explained intuitively by the fact that max-pooling relies merely on the peak of a function (or a region of interest), whereas in our model, pooling is defined collectively based on a number of elements near the maximum. Consequently, those peaks that are outliers and likely caused by noise get normalised by other pooled elements. The rationale within our model is to pool a larger percentage at low contrast since in those conditions, peaks are not informative on their own, whereas at high contrast peaks are likely to be more informative and other irrelevant details must be removed (therefore a smaller percentage is pooled).

Although, the importance of choosing an appropriate pooling type has been demonstrated both experimentally~\cite{jarrett2009best,yang2009linear}, and theoretically~\cite{boureau2010theoretical}, current standard pooling mechanisms lack the desired generalisation~\cite{murray2014generalized}. We believe that contrast-variant-pooling can offer a more dynamic and general solution. In this article, we evaluated CVP on the colour constancy phenomenon as a proof-of-concept, however our formulation of CVP is generic (and based on local contrast) and in principle it can be applied to a wider range of computer vision algorithms, such as deep-learning, where pooling is a decisive factor~\cite{scherer2010evaluation}.

In our implementation of CVP, we approximated local contrast through local standard deviation (see Eq.~\ref{eq:contrastcalculation}). There are at least two factors that require a more profound analysis: (i) incorporating more sophisticated models of contrast perception~\cite{haun2013perceived} accounting for extrema in human contrast sensitivity; and (ii) analysing the role of kernel size in the computation of local contrast.

\section{Conclusion}
\label{sec:conclusion}

In this article, we presented a novel biologically-inspired contrast-variant-pooling (CVP) mechanism grounded in the physiology of the visual cortex. Our main contribution can be summarised as linking the percentage of pooled signal to the local contrast of the stimuli, \ie pooling a larger percentage at low contrast and a smaller percentage at high contrast. Our CVP operator remains always closer to max-pooling rather than to sum-pooling since natural images generally contain more homogeneous areas than abrupt discontinuities.

We tested the efficiency of our CVP model in the context of colour constancy by replacing the max-pooling operator of four algorithms with the proposed pooling. After that, we conducted experiments in three benchmark datasets. Our results show that contrast-variant-pooling outperforms the commonly used max-pooling operator nearly in all cases. This can be explained by the fact that our model allows for more informative peaks to be pooled while suppressing less informative peaks and outliers.

We further argued that the proposed CVP is a generic operator, thus its application can be extended to a wider range of computer vision algorithms by offering a dynamic (and automatic) framework that is based on the local contrast of an image or a pixel. This opens a multitude of possibilities for future lines of research and it remains an open question whether our model can reproduce its excellent results in other domains as well. Therefore, it certainly is interesting to investigate whether our CVP can improve convolutional neural networks.

\section*{Acknowledgements}

This work was funded by the Spanish Secretary of Research and Innovation (TIN2013-41751-P and TIN2013-49982-EXP).

\bibliography{egbib}
\end{document}